\documentclass{article}

\PassOptionsToPackage{numbers, compress}{natbib}

\usepackage[final]{neurips_2021}

\usepackage[utf8]{inputenc} 
\usepackage[T1]{fontenc}    
\usepackage{url}            
\usepackage{booktabs}       
\usepackage{amsfonts}       
\usepackage{nicefrac}       
\usepackage{microtype}      
\usepackage{xcolor}         
\usepackage{graphicx}
\usepackage{wrapfig}
\usepackage{multirow}
\usepackage[font=small]{caption}

\usepackage{amsmath}

\usepackage{amssymb}

\usepackage[pagebackref=true,breaklinks=true,letterpaper=true,colorlinks,bookmarks=false]{hyperref}

\usepackage{cleveref} 
\crefname{figure}{fig.}{figs.} 
\Crefname{figure}{Fig.}{Figs.} 

\title{XC: Exploring Quantitative Use Cases for Explanations in 3D Object Detection}

\author{%
  Sunsheng Gu, \hspace{0.5cm} Vahdat Abdelzad, \hspace{0.5cm} Krzysztof Czarnecki \\
  University of Waterloo\\
  \texttt{ssgu@uwaterloo.ca} \\
  \texttt{vahdat.abdelzad@uwaterloo.ca} \\
  \texttt{krzysztof.czarnecki@uwaterloo.ca} \\
}

\begin{document}
\setcitestyle{square}
\setcitestyle{citesep={,}}
\maketitle

\begin{abstract}
Explainable AI (XAI) methods are frequently applied to obtain qualitative insights about deep models' predictions. However, such insights need to be interpreted by a human observer to be useful. In this paper, we aim to use explanations directly to make decisions without human observers. We adopt two gradient-based explanation methods, Integrated Gradients (IG) and backprop, for the task of 3D object detection. Then, we propose a set of quantitative measures, named Explanation Concentration (XC) scores, that can be used for downstream tasks. These scores quantify the concentration of attributions within the boundaries of detected objects. We evaluate the effectiveness of XC scores via the task of distinguishing true positive (TP) and false positive (FP) detected objects in the KITTI and Waymo datasets. The results demonstrate an improvement of more than 100\% on both datasets compared to other heuristics such as random guesses and the number of LiDAR points in the bounding box, raising confidence in XC's potential for application in more use cases. Our results also indicate that computationally expensive XAI methods like IG may not be more valuable when used quantitatively compare to simpler methods.
\end{abstract}

\section{Introduction}
\label{s:intro}

\begin{wrapfigure}{r}{0.45\columnwidth}
\centering
\raisebox{0pt}[\dimexpr\height-2\baselineskip\relax]{
\includegraphics[width=0.45\textwidth]{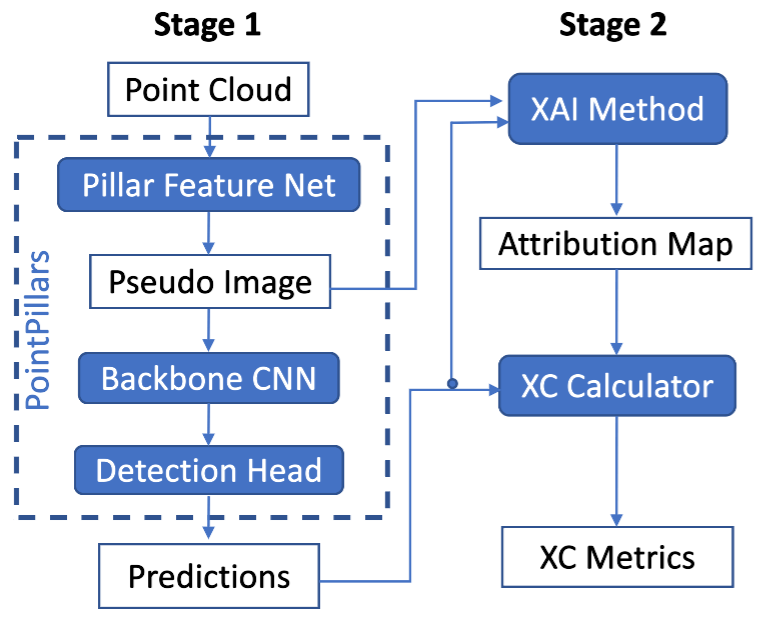}
}%
\caption{Overview of the XC calculation process: PointPillars first process the input point cloud once, then an XAI method computes feature attribution map for specific predictions using the pseudo image as input, and XC metrics are obtained.}
\label{f:XC_flowchart}
\vspace{-15pt}
\end{wrapfigure}

Recent development in deep neural networks (DNNs) has led to the state of the art performance on 2D \cite{Girshick2013,He2015,Liu16,Redmon2015,Ren2015} and 3D \cite{Lang_2019_CVPR,Shi_2019_CVPR,Zhou_2018_CVPR} object detection tasks. However, despite of the improvement in model performance, the lack of model interpretability remains a significant drawback. This issue has sparked interest in \emph{explainable artificial intelligence} (XAI). The primary goal of these XAI methods is to uncover the logic behind the models' decisions \cite{LiXAI2020}.

Several recent methods have been proposed to generate interpretable visual explanations \cite{simonyan2014deep, zeiler2013visualizing, springenberg2015striving, Selvaraju_2017_ICCV, IG_2017_ICML}. They are usually analyzed by a human observer to understand the model's reasoning for some particular decisions. Hence, explanations are very useful for debugging and diagnosis. However, it would be impossible for a human to analyze each instance in a large dataset. To analyze explanations on a large scale, it is necessary to derive quantitative measures from the explanations. There have been a few works in this direction \cite{ijcai2017-371, erion2020improving, rieger2020interpretations}, but the input data experimented on are 2D image, text, or tabular data. We have not found any experiments on explanations-related quantitative measures applied to LiDAR point cloud input or to the task of object detection. 
Our motivation is to fill in the gap by exploring quantitative usage of explanations for LiDAR-based object detection.

We study explanations in the form of attribution maps for a 3D object detector named PointPillars \cite{Lang_2019_CVPR} and use them for the problem of distinguishing true positive (TP) vs. false positive (FP) predictions. Attributions, in the context of XAI, denote the influence of input features on model output. We generate attribution maps using two XAI methods: 1) Integrated Gradients (IG) \cite{IG_2017_ICML}, selected for its axiomatic properties; 2) backpropagation \cite{simonyan2014deep}, selected for its low computational cost.
Our main contributions are \footnote{Our code is available at \url{https://github.com/SunshengGu/XC_eval_pcdet}.}:
\begin{itemize}
    \item We demonstrate that quantitative usage of explanation for 3D object detection is a promising direction for future research: We propose a set of quantitative metrics called \emph{Explanation Concentration} (XC), which measures the concentration of attributions within each predicted object. XC can be used to classify the TP vs. FP predictions effectively, often achieving more than 100\% improvement compared to simple heuristics such as number of LiDAR points within a predicted box. 
    \item We discover that XC scores derived from backpropagation can perform better than those derived from IG.
    \item We propose a new score that can identify TP vs. FP pedestrian predictions better than the individual XC scores and object class score. This score is generated by combining the XC scores with object class score using a MLP.
\end{itemize}

\section{Related work}
\label{s:related}
\noindent
\textbf{Explanation methods}
A simple form of explanation is an input feature saliency map obtained via backpropagation: one computes the partial derivatives of the model output with respect to the input features \cite{simonyan2014deep}. These partial derivatives are a measure of importance (also called \emph{attributions}) for the input features. Deconvolution \cite{zeiler2013visualizing} and Guided-backprop \cite{springenberg2015striving} are similar methods. 
Integrated Gradients (IG) \cite{IG_2017_ICML} is another gradient-based explanation method that computes multiple inputs along a straight-line path from a baseline input to the original input through affine transformations. IG then combines gradients from all these inputs to get feature attributions. The baseline input is defined as an input which generates zero output. In the case of image input, the baseline is defined as a black image with pixel values all equals to zero. Sundararajan \emph{et al.} \cite{IG_2017_ICML} demonstrate that IG satisfies several axiomatic properties for explanations such as sensitivity (attribution values for nonzero features are nonzero, and attribution values for zero-valued features are zero) and completeness (sum of attributions equals to model output value). 

\noindent
\textbf{Quantitative usage of explanations}
Several attempts have been made to utilize explanations quantitatively. Ross \emph{et al.} \cite{ijcai2017-371} use the gradient of the log of model output with respect to input features as attributions. They design a loss term using prior knowledge about feature relevance to punish the model for having attributions for irrelevant features, hence forcing the model to be "right for the right reasons" by focusing on the relevant features only. Rieger \emph{et al.} \cite{rieger2020interpretations} and Du \emph{et al.} \cite{crex2019} also propose similar methods to constrain feature attribution values. Erion \emph{et al.} \cite{erion2020improving} realize that such prior knowledge about input feature relevance would not always be available, hence they propose a new loss named ``pixel attribution prior" (PAP) which only aims to reduce variations in attribution values for neighbouring pixels. Their reasoning is that a robust model should not be heavily influenced by small variations in a few disconnected pixels; rather, it should learn patterns from patches of pixels and neighbouring pixels should have similar attributions.

\noindent
\textbf{LiDAR-based object detection}
A popular technique for point cloud-based object detection is to first divide the 3D point cloud into grids called voxels, learn features for each voxel, then apply convolutional layers to the voxel-wise features to extract higher level features. Techniques used for 2D object detection, such as region proposals and anchor boxes, can then be applied to the extracted feature maps. VoxelNet \cite{Zhou_2018_CVPR}, SeCOND \cite{yan2018second}, and PointPillars \cite{Lang_2019_CVPR} all partly adopted this strategy. PointPillars' fast inference speed (62 Hz) makes it a desirable choice for deployment on embedded hardware, such as in autonomous vehicles.

\noindent
\textbf{False positive detection}
Several studies have been conducted on identifying false positive (FP) predictions. Hendrycks and Gimpel \cite{Hendrycks2017} demonstrated that the softmax class probability could effectively distinguish true positive (TP) and false positive (FP) predictions in multiple datasets, with performance far exceeding random guess. Chen \emph{et al.} \cite{chen2021box} also used class score to filter out FP pedestrians predictions iteratively. 
Methods designed for identifying adversarial attacks or detecting out-of-distribution (OOD) samples may also be applied to detect FP samples. Some notable works are feature squeezing \cite{Xu2018squeeze}, LID \cite{Ma2018LID}, GraN \cite{Lust2020GraN}, ODIN \cite{Liang2018ODIN} and Lee \emph{et al.} \cite{Lee2018Mahala}.


\begin{figure*}[htp]
\centering
\includegraphics[width=\textwidth]{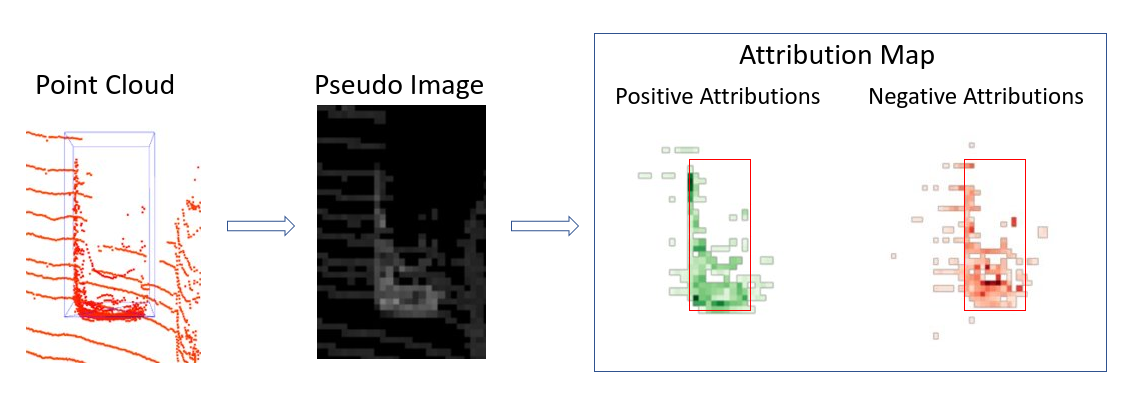}
\caption{Cropped IG attribution map visualization for a car prediction. The blue box in the point cloud is the bounding box for the ground truth object. The red box in the attribution map is the 2D projection of the car prediction bounding box. Positive attributions are indicated by green pixels and negative attributions are indicated by red pixels; darker color means greater magnitude. Pixels with attribution values less than 0.1 are filtered out and appear white. Note that the attribution map has the same resolution as the pseudo image.}
\label{f:attr_viz}
\end{figure*}

\section{Explanation in the form of attributions}
\label{s:attr}

As mentioned before in \Cref{s:related}, explanation can be presented in the form of input feature attributions, where the magnitude of the attributions corresponds to feature importance, and the sign of attributions indicates positive or negative influence on the model output. In our approach, attributions are generated by IG \cite{IG_2017_ICML} and backprop \cite{simonyan2014deep} for PointPillars \cite{Lang_2019_CVPR}. PointPillars has three parts as shown in \Cref{f:XC_flowchart}: a pillar feature net which learns a 2D bird's eye view (BEV) voxelized feature map called the \emph{psuedo image} from the original point cloud, a backbone 2D CNN to extract more features from the psuedo image, and SSD \cite{Liu16} as detection head to generate 3D object predictions.  

IG's effectiveness on image data is well-recognized.
However, there are a fixed number of pixels at fixed locations in an image; whereas in a point cloud, both the quantity and location of the points can vary greatly. IG's input transformations may not be meaningful for point cloud. Hence, we chose to generate attributions using the pseudo image as input, so that IG could be directly applied. Note that our pipeline in \Cref{f:XC_flowchart} is not limited to IG or backprop explanations only, any other XAI method that produces explanations in the form of feature attributions could be applied too.

There could be many predictions in the same pseudo image. To explain the different predictions separately, each attribution map is generated for a specific predicted box and its class label. In \Cref{f:attr_viz}, the IG attributions generated for a car prediction is shown. For this particular example, the positive attributions highlight features that increase the model's belief that the object is a car, whereas the negative attributions reduce this belief. 

\section{Explanation concentration (XC)}
\label{s:xc_calc}

We propose a new set of scores called \emph{Explanation Concentration} (XC) which measures the concentration of attributions within a given predicted object's boundaries. As shown in \Cref{f:attr_viz}, many pixels outside of the object have noticeable attributions too, indicating that context features can influence model output as well. The XC scores are partly motivated by previous studies which demonstrate that overly-relying on context can hurt model performance for both image classification \cite{Shetty2018} and 3D object detection \cite{Shi_2019_CVPR}.

The process of computing the XC scores is as follows. First, denote the $i^{th}$ predicted box in a pseudo image as $Pred_i$. Then denote the sum of positive attributions across all channels in the pixel at location $(x,y)$ of the attribution map as $a^{+}_{(x,y)}$. The idea is to avoid having positive and negative channel values cancelling each other out at a specific pixel. 

Using thresholds to eliminate noisy signals is common practice in computer vision. For example, in the Canny edge detection algorithm \cite{canny1986computational}, thresholds are used to mask out weak edges. We apply the same idea to the attribution values. Denote a pixel-wise threshold as $a_{thresh}$, and use it to filter out insignificant attributions (in other words, the sum $a^{+}_{(x,y)}$ can be called \emph{significant} if it exceeds $a_{thresh}$). An indicator function $I^{+}_{(x,y)}$ is defined such that it equals to $1$ if $a^{+}_{(x,y)} >= a_{thresh}$, $0$ otherwise. One way to quantify the concentration of attributions within the predicted box $Pred_i$ is by summing. Two new variables are defined for each $Pred_i$ in a pseudo image:
\begin{equation}
s^{+}_i = \sum{a^{+}_{(x,y)} \times I^{+}_{(x,y)}},\ \forall (x,y)\ \mathrm{in\ } Pred_i
\label{eq: a_sum_in}
\end{equation}
\begin{equation}
S^{+}_i = \sum{a^{+}_{(x,y)} \times I^{+}_{(x,y)}},\ \forall (x,y)\ \mathrm{in\ pseudo \; image}
\label{eq: a_sum_all}
\end{equation}
Now the stage is set for defining an XC score by summing:
\begin{equation}
XC\_s^{+}_i = s^{+}_i / S^{+}_i
\label{eq: XC_sum}
\end{equation}
$XC\_s^{+}_i$ is thus the proportion of the positive attributions which lie within $Pred_i$. 

The other way to quantify the concentration of attributions for $Pred_i$ is by counting. For this purpose, we count the number of pixels in the pseudo image having significant attributions, rather than summing them:
\begin{equation}
c^{+}_i = \sum{I^{+}_{(x,y)}},\ \forall (x,y)\ \mathrm{in\ } Pred_i
\label{eq: a_cnt_in}
\end{equation}
\begin{equation}
C^{+}_i = \sum{I^{+}_{(x,y)}},\ \forall (x,y)\ \mathrm{in\ pseudo image}
\label{eq: a_cnt_all}
\end{equation}
\begin{equation}
XC\_c^{+}_i = c^{+}_i / C^{+}_i
\label{eq: XC_cnt}
\end{equation}

\begin{wrapfigure}{r}{0.4\columnwidth}
\centering
\raisebox{0pt}[\dimexpr\height-1.2\baselineskip\relax]{
\includegraphics[width=0.4\textwidth]{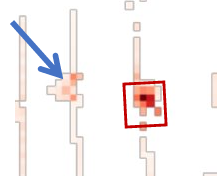}
}%
\caption{Negative attribution map for a TP pedestrian prediction (the red box). The patch of outlier attributions is pointed out by the arrow. }
\label{f:XC_outlier}
\vspace{-15pt}
\end{wrapfigure}

Computing XC by counting might be helpful, because often there are a few outlier pixels with high magnitude attribution values outside of the bounding bounding box (see \Cref{f:XC_outlier} for example). When XC is computed by counting, these outliers will not have a big effect; but when XC is computed by summing, the values of these outlier pixels can skew the resulting XC value towards the lower end. One may also compute a similar set of scores by considering the negative attributions only: call them $XC\_s^{-}_i$ and $XC\_c^{-}_i$ for a certain box $Pred_i$. 

We observed that pixels located at object boundaries often get labelled as ``outside of the box''. Hence, when calculating any XC scores, the predicted boxes are enlarged by a small margin $m$ on all sides, so that pixels at object boundaries are labeled as inside the predicted box.

\section{Results and discussions}
\label{s:results}

\subsection{Evaluation metrics and implementation details}
\label{ss:metrics_settings}
The objective of our experiment is to evaluate XC's performance on a meta classification task: classifying predictions as either FP or TP. The predicted objects are categorized as TP or FP based on KITTI's conventions \cite{Geiger2012CVPR}. To evaluate the performance of a specific score, we could simply apply a score threshold: if the score is above the threshold then the corresponding prediction is TP, otherwise it's FP. Then we can evaluate the resulting detection accuracy. However, the resulting accuracy is a function of the threshold. To remove the effect of threshold selection and evaluate the performance of different XC scores more fairly, we compute area under the precision recall curve (AUPR) \cite{Manning1999} and area under the receiver operating characteristics curve (AUROC) \cite{Davis2006}, both are threshold-independent performance measures for binary classification.

In binary classification tasks, typically one class is treated as the ``positive'' class, whereas the other class is treated as the ``negative'' class. We may choose to treat either the TP boxes or the FP boxes as the positive class. The AUROC metric treats both classes equally and can reflect the score's (e.g., one of the XC scores) ability in correctly identifying both the positive and negative classes. On the other hand, the AUPR metric puts more emphasis on the score's ability to correctly identify the positive class. For both AURP and AUROC, higher values indicate better performance. 

We evaluate the XC scores on three PointPillars \cite{Lang_2019_CVPR} models trained on the KITTI dataset \cite{Geiger2012CVPR} and on one PointPillars model trained on the Waymo dataset \cite{waymo2020}. We use OpenPCDet's \cite{openpcdet2020} implementation of PointPillars for our experiments and adapt their default settings for training. The first three models are trained for 80 epochs on the KITTI dataset, with 3712 frames for training and 3769 frames for validation. Due to time and resource constraints, we obtain only one more model trained for 30 epochs on 20\% of the Waymo dataset, with 31616 frames for training and 7997 frames for validation. To obtain attribution values, we use Captum \cite{captum2020}, a model interpretability library developed for PyTorch \cite{NEURIPS2019_9015}. We explore both IG \cite{IG_2017_ICML} and backprop \cite{simonyan2014deep} as explanation methods. 

For KITTI, the XC scores are obtained on the predicted objects from the 3769 validation frames. There are on average 67k predicted objects produced by each model. One pixel in the pseudo image encodes point features in a $0.16$m $\times 0.16$m $\times 4.00$m pillar. For Waymo, we comptue XC scores for 94k predicted boxes sampled from 800 validation frames. One pixel in the pseudo image represents point features in a $0.33$m $\times 0.33$m $\times 6.00$m pillar. For both datasets, we apply $m = 0.2$m to the predicted boxes and apply $a_{thresh} = 0.1$ to the attribution maps.



\begin{figure}[htp]
\centering
\includegraphics[width=\columnwidth]{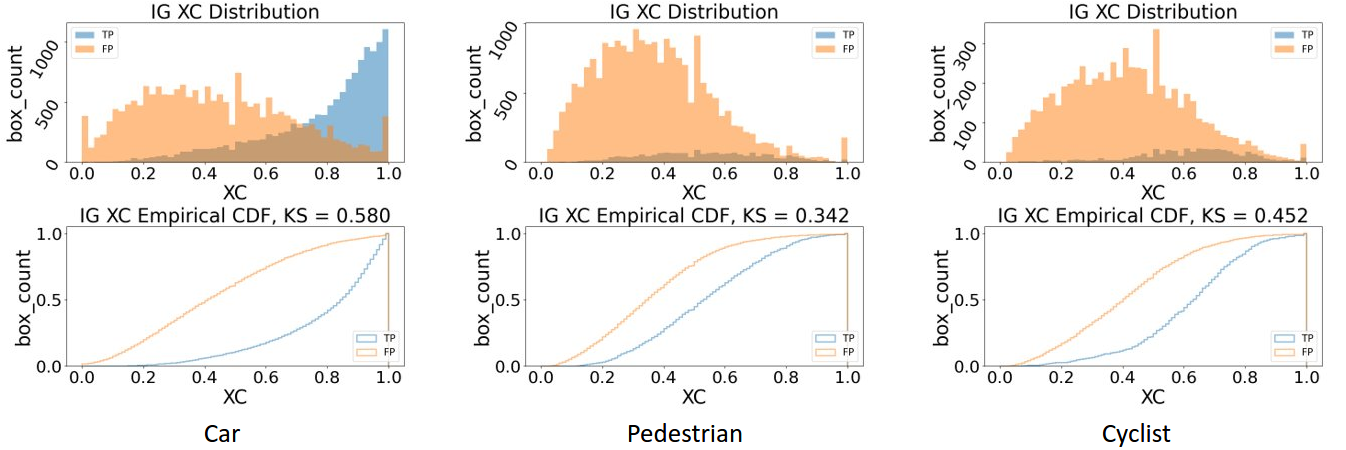}

\vspace{0.2cm}

\includegraphics[width=\columnwidth]{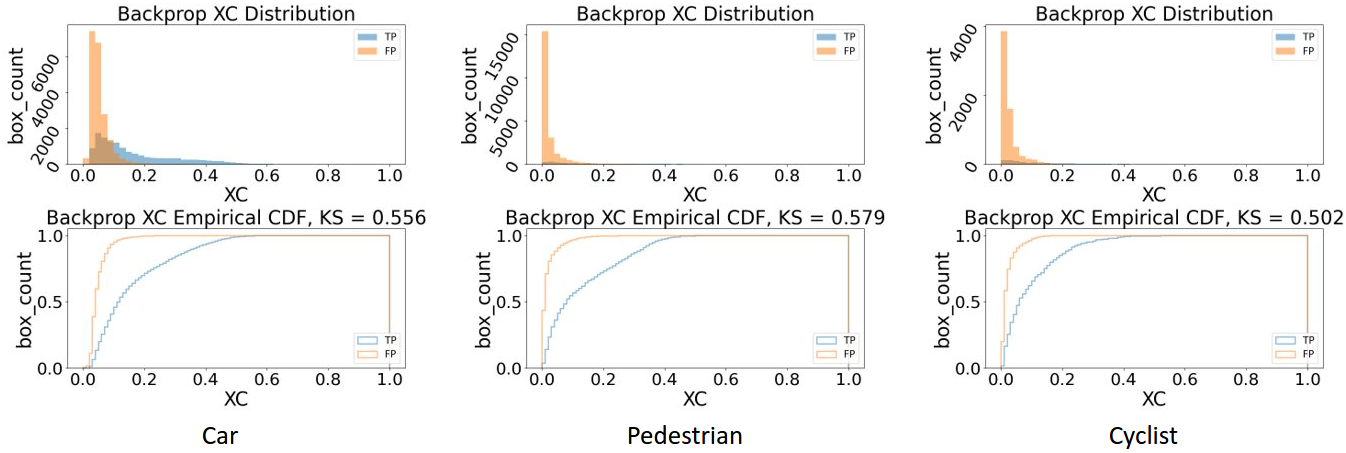}
\caption{Distribution of $XC\_c^{+}$ values obtained by IG (upper row) and backprop (lower row) attributions for TP and FP predicted boxes in each of the object class of the KITTI validation set. ``KS'' means the Kolmogorov-Smirnov statistic \cite{KS2008}. }
\label{f:xc_dist}
\end{figure}

\subsection{Distribution of XC values}
\label{ss:xc_distr}

The distribution of $XC\_c^{+}$ values obtained from one of the models trained on KITTI \cite{Geiger2012CVPR} is shown in \Cref{f:xc_dist}. Both the histograms and the empirical cumulative distribution function plots demonstrate that the XC values of TP instances tend to be greater than those of the FP instances. Another notable observation is that XC values derived from IG are well-dispersed within the range of $[0, 1]$ for both the TP and FP instances, whereas XC values derived from backprop are mostly below 0.5, with almost all FP instances having XC values below 0.2.


\subsection{Using XC to identify TP and FP predictions}
\label{ss:xc_detect_tp_fp}
The results for TP vs. FP box classification using the XC scores on the KITTI dataset \cite{Geiger2012CVPR} are shown in \Cref{t:XC_compare_kitti}. Each specific metric in this table is averaged over three models. The four XC scores derived from IG and backprop are evaluated, along with four other box-wise features (random guess, distance of the predicted box to LiDAR sensor, the number of points inside the predicted box, and the highest class score for the predicted box) serving as baselines for comparison. Each of the aforementioned features are evaluated by three metrics on different object classes, making up twelve metrics in total. Note that the object classes in \Cref{t:XC_compare_kitti} (as well as those in other tables in later sections) represent the predicted labels, not the ground truth labels. Also, in OpenPCDet's \cite{openpcdet2020} implementation of PointPillars \cite{Lang_2019_CVPR}, the class scores are not softmax scores. Rather, they are class-wise sigmoid scores: a pair of ``class vs. not class'' scores for each object class. 

When computing AUROC and AUPR, the TP boxes are treated as the positive instances, but AUPR\_op refers to the AUPR value obtained by treating the FP predicted boxes as positive instances. In essence, AUPR reflects TP detection performance, whereas AUPR\_op reflects FP detection performance. The expected AUROC value for a random score is 0.5, and the expected AUPR value is the proportion of the positive instances. 

\begin{table*}[htp]
\caption{Comparison of the XC scores' ability to classify TP and FP predictions for different object types in the KITTI dataset. The subscripts ``IG'' and ``B'' indicate whether the corresponding XC score is derived from IG or backprop attributions. For each evaluation metric, the XC scores performing worse than number of points are highlighted by \underline{underscore} and the best performing feature other than the top class score is highlighted in \textbf{bold}.}

\begin{center}
\resizebox{\textwidth}{!}{%
\begin{tabular}{ c|c|c|c|cccc|cccc|c } 
\toprule
  Metrics & Random & Distance & Points & $XC\_s^{+}_{IG}$ & $XC\_c^{+}_{IG}$ &$XC\_s^{-}_{IG}$ & $XC\_c^{-}_{IG}$ & $XC\_s^{+}_{B}$ & $XC\_c^{+}_{B}$ &$XC\_s^{-}_{B}$ & $XC\_c^{-}_{B}$ & Top Class Score \\
  \midrule
  \shortstack{\textbf{AUROC}\\All} & 0.5 & 0.669 & 0.720 & 0.843 & 0.868 & 0.827 & 0.869 & 0.823 & 0.903 & 0.822 & \textbf{0.908} & 0.971 \\
  Car & 0.5 & 0.770 & 0.779 & 0.837 & 0.857 & 0.822 & 0.857 & \underline{0.708} & 0.861 & \underline{0.698} & \textbf{0.869} & 0.964 \\
  Pedestrian & 0.5 & 0.779 & 0.832 & \underline{0.648} & \underline{0.699} & \underline{0.671} & \underline{0.754} & 0.864 & 0.888 & 0.882 & \textbf{0.894} & 0.958 \\
  Cyclist & 0.5 & 0.635 & 0.780 & 0.810 & 0.797 & 0.806 & 0.824 & 0.808 & 0.843 & 0.819 & \textbf{0.855} & 0.965 \\
  \midrule
  \shortstack{\textbf{AUPR}\\All} & 0.232 & 0.372 & 0.464 & 0.585 & 0.653 & 0.551 & 0.635 & 0.597 & 0.786 & 0.577 & \textbf{0.794} & 0.926 \\
  Car & 0.391 & 0.636 & 0.666 & 0.713 & 0.749 & 0.697 & 0.747 & \underline{0.621} & 0.830 & \underline{0.597} & \textbf{0.836} & 0.948 \\
  Pedestrian & 0.071 & 0.215 & 0.250 & \underline{0.116} & \underline{0.153} & \underline{0.136} & \underline{0.190} & 0.518 & 0.557 & 0.541 & \textbf{0.572} & 0.759 \\
  Cyclist & 0.083 & 0.167 & 0.203 & 0.225 & 0.237 & 0.239 & 0.270 & 0.425 & 0.534 & 0.450 & \textbf{0.554} & 0.829 \\
  \midrule
  \shortstack{\textbf{AUPR\_{op}}\\All} & 0.768 & 0.859 & 0.878 & 0.941 & 0.950 & 0.937 & 0.952 & 0.936 & 0.965 & 0.937 & \textbf{0.967} & 0.989 \\ 
  Car & 0.609 & 0.829 & 0.839 & 0.894 & 0.903 & 0.885 & \textbf{0.905} & \underline{0.761} & 0.888 & \underline{0.761} & 0.897 & 0.974 \\
  Pedestrian & 0.929 & 0.976 & 0.984 & \underline{0.959} & \underline{0.967} & \underline{0.962} & \underline{0.973} & 0.986 & 0.989 & 0.988 & \textbf{0.990} & 0.996 \\
  Cyclist & 0.917 & 0.939 & 0.976 & 0.977 & \underline{0.975} & 0.978 & 0.980 & \underline{0.975} & 0.979 & 0.977 & \textbf{0.982} & 0.992 \\
  \bottomrule
\end{tabular}}
\end{center}
\label{t:XC_compare_kitti}
\end{table*}


Referring to \Cref{t:XC_compare_kitti}, it is clear that the top class score beats all other features in every evaluation metric. However, the improvement brought by XC is certainly non-trivial. One can observe that for all features evaluated, the second best performing feature is most often $XC\_c^{-}_B$. Although distance to sensor and number of points in predicted box beat the XC scores generated by IG on the three pedestrian class metrics, they are unable to beat any of the XC scores generated by backprop on the pedestrian class. 

A very notable case is the AUPR for pedestrian predictions: $XC\_c^{-}_{B}$ achieves 129\% improvement compared to number of points and 706\% improvement compare to random guess. The other three XC scores derived from backprop also achieve more than 100\% on AUPR for pedestrian compared to the number of points. Improvements of similar magnitude are also achieved by the backprop XC scores on AUPR for the cyclist class. These observations indicate that the backprop XC scores are much better at correctly identifying the TP predictions for the pedestrian and cyclist classes than simple heuristics such as number of points. Another interesting observation is that the XC scores derived by counting usually outperforms those derived by summing. Such improvement might be due to the outlier attenuation effect mentioned in the second last paragraph of \Cref{s:xc_calc}. 

\begin{table}[htp]
\caption{Comparison of the XC scores' ability to classify TP and FP predictions for different object types in the Waymo dataset. For each evaluation metric, the XC scores performing worse than number of points are highlighted by \underline{underscore} and the best performing feature other than the top class score is highlighted in \textbf{bold}.}
\begin{center}
\resizebox{\textwidth}{!}{%
\begin{tabular}{ c|c|c|c|cccc|cccc|c } 
\toprule
  Metrics & Random & Distance & Points & $XC\_s^{+}_{IG}$ & $XC\_c^{+}_{IG}$ &$XC\_s^{-}_{IG}$ & $XC\_c^{-}_{IG}$ & $XC\_s^{+}_{B}$ & $XC\_c^{+}_{B}$ &$XC\_s^{-}_{B}$ & $XC\_c^{-}_{B}$ & Top Class Score \\
  \midrule
  \shortstack{\textbf{AUROC}\\All} & 0.5 & 0.609 & 0.701 & 0.714 & 0.758 & 0.738 & 0.766 & 0.729 & 0.788 & 0.721 & \textbf{0.793} & 0.965 \\ 
  Vehicle & 0.5 & 0.703 & 0.809 & \underline{0.799} & 0.821 & \underline{0.806} & 0.823 & \underline{0.754} & \textbf{0.860} & \underline{0.725} & \textbf{0.860} & 0.982 \\
  Pedestrian & 0.5 & 0.528 & 0.614 & \underline{0.529} & \underline{0.529} & \underline{0.529} & \underline{0.545} & \underline{0.605} & 0.638 & 0.646 & \textbf{0.651} & 0.927 \\
  Cyclist & 0.5 & 0.627 & 0.738 & 0.766 & \underline{0.725} & \underline{0.673} & \underline{0.689} & 0.794 & 0.823 & 0.799 & \textbf{0.823} & 0.979 \\
  \midrule
  \shortstack{\textbf{AUPR}\\All} & 0.276 & 0.343 & 0.476 & \underline{0.460} & 0.535 & 0.483 & 0.540 & 0.569 & \textbf{0.700} & 0.542 & 0.699 & 0.936 \\ 
  Vehicle & 0.377 & 0.595 & 0.721 & \underline{0.626} & \underline{0.659} & \underline{0.624} & \underline{0.654} & \underline{0.678} & \textbf{0.827} & \underline{0.648} & 0.824 & 0.973 \\
  Pedestrian & 0.176 & 0.121 & 0.147 & 0.183 & 0.184 & 0.181 & 0.192 & 0.284 & 0.323 & 0.326 & \textbf{0.340} & 0.829 \\
  Cyclist & 0.051 & 0.072 & 0.113 & \underline{0.111} & \underline{0.096} & \underline{0.078} & \underline{0.082} & 0.279 & 0.377 & 0.283 & \textbf{0.394} & 0.791 \\
  \midrule
  \shortstack{\textbf{AUPR\_{op}}\\All} & 0.724 & 0.811 & 0.862 & 0.863 & 0.881 & 0.874 & \textbf{0.884} & \underline{0.850} & 0.876 & \underline{0.857} & 0.882 & 0.983 \\ 
  Vehicle & 0.623 & 0.784 & 0.873 & 0.878 & 0.891 & 0.883 & 0.892 & \underline{0.809} & 0.891 & \underline{0.786} & \textbf{0.893} & 0.987 \\
  Pedestrian & 0.824 & 0.897 & \textbf{0.925} & \underline{0.841} & \underline{0.839} & \underline{0.842} & \underline{0.846} & \underline{0.866} & \underline{0.882} & \underline{0.889} & \underline{0.888} & 0.979 \\
  Cyclist & 0.950 & 0.967 & 0.984 & 0.984 & \underline{0.980} & \underline{0.976} & \underline{0.978} & 0.984 & 0.986 & 0.985 & \textbf{0.986} & 0.999 \\
  \bottomrule
\end{tabular}}
\end{center}
\label{t:XC_compare_waymo}
\end{table}

To ensure that the advantages offered by XC are not specific to the KITTI dataset \cite{Geiger2012CVPR}, we also present results from Waymo dataset \cite{waymo2020} in \Cref{t:XC_compare_waymo}. Again, for most evaluation metrics, one of the XC scores is the second best performing feature besides the top class score. 
In addition, the XC scores obtained from backprop often show 100\% or more improvement on AUPR for pedestrian and cyclist predictions compared to the number of points. And again, $XC\_c^{-}_B$ is the best performing feature in most cases. Hence, we believe that the advantages of XC are not limited to the KITTI dataset only. 

\begin{wraptable}{r}{6cm}
\vspace{-30pt}
\caption{Average XC performance on distinguishing TP vs. FP predictions for the KITTI dataset. 
The highest value in each row is highlighted in \textbf{bold}.}
\begin{center}
\raisebox{0pt}[\dimexpr\height-0.5\baselineskip\relax]{
\resizebox{0.41\columnwidth}{!}{%
\begin{tabular}{ c|c|c|c } 
\toprule
  Metrics & $XC_{IG}$ & Modified $XC_{IG}$ & $XC_{B}$ \\
  \midrule
  \shortstack{\textbf{AUROC}\\All}  & 0.852 & 0.861 & \textbf{0.864} \\ 
  Vehicle & \textbf{0.843} & 0.795 & 0.784 \\
  Pedestrian  & 0.693 & 0.848 & \textbf{0.882} \\
  Cyclist  & 0.809 & 0.812 & \textbf{0.831} \\
  \midrule
  \shortstack{\textbf{AUPR}\\All} & 0.606 & 0.657 & \textbf{0.689} \\ 
  Vehicle & \textbf{0.726} & 0.707 & 0.721  \\
  Pedestrian & 0.149 & 0.321 & \textbf{0.547} \\
  Cyclist & 0.243 & 0.282 & \textbf{0.490} \\
  \midrule
  \shortstack{\textbf{AUPR\_{op}}\\All} & 0.945 & \textbf{0.952} & 0.951 \\ 
  Vehicle & \textbf{0.897} & 0.848 & 0.827 \\
  Pedestrian & 0.965 & 0.986 & \textbf{0.988}\\
  Cyclist & 0.977 &\textbf{0.978} & \textbf{0.978} \\
  \bottomrule
\end{tabular}}}%
\end{center}
\label{t:avg_xc_performance}
\vspace{-20pt}
\end{wraptable}

\subsection{Why backprop outperforms IG?}
\label{ss:backprop_better}

IG \cite{IG_2017_ICML} is designed to reflect feature importance more precisely than other simpler XAI methods such as backprop. Thus, it would be interesting to know why IG-based XC scores underperform backprop-based XC scores in the task of classifying TP vs. FP predictions, especially for the pedestrian and cyclist predictions (see AUPR in \Cref{t:XC_compare_kitti} and \Cref{t:XC_compare_waymo}). We suspect that this is due to the difference in XC distribution. In \Cref{f:xc_dist}, we present the KS statistic \cite{KS2008} between the distributions of XC values in the TP and FP instances of each object class. The KS statistic is a measure for goodness of fit between two distributions: greater value indicates greater difference between the two distributions. Note that the backprop-based XC scores are able to produce much KS statistic between TP and FP distributions in the pedestrian and cyclist class than the IG-based scores. 

To alter the distribution of IG-based XC scores, we remove the last step in computing IG attributions. As mentioned before in \Cref{s:related}, IG zeros out attributions for input features with zero value. This is often achieved by multiplying the computed attributions at each input location $i$ by $(x_{i} - x_{i}^\prime)$, where $x$ is the input and $x^\prime$ is the baseline, which is by default set to zero. We remove this process and re-calculate IG-derived XC scores on the KITTI dataset \cite{Geiger2012CVPR}. Then we evaluate the new XC scores on the binary classification task for TP vs. FP predictions. The average results of all four IG-derived XC scores, all four backprop-derived XC scores (these are averaged over the values presented in \Cref{t:XC_compare_kitti}), and of the four modified IG-derived XC scores are presented in \Cref{t:avg_xc_performance}. Note that the results are also averaged over all three models trained on KITTI. The modified IG XC scores resulted in 115\% improvement in AUPR for pedestrian predictions and in 16\% improvement in AUPR for cyclist predictions compared to the original IG XC scores. Improvement can also be observed in AUROC and AUPR\_op for the pedestrian and cyclist classes. 

\begin{figure}[htp]
\centering
\includegraphics[width=\columnwidth]{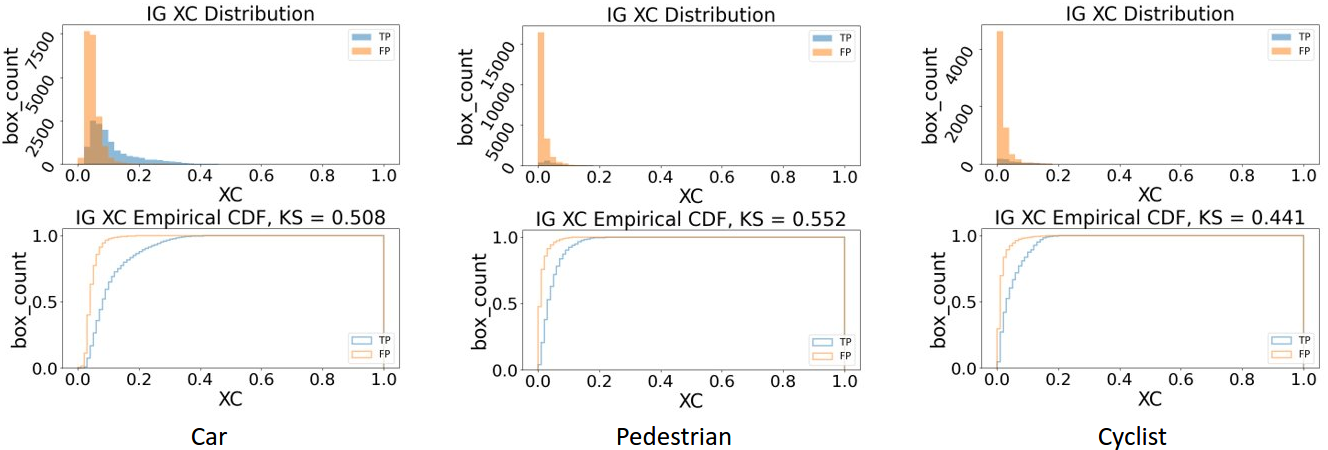}
\caption{Distribution of $XC\_c^{+}$ values obtained by modified IG (without multiplying attributions by input) attributions for TP and FP predicted boxes in the KITTI validation set.}
\label{f:xc_dist_ig_no_mult}
\end{figure}

The improvements on pedestrian and cyclist objects also coincide with a shift in the distribution of XC values. As shown in \Cref{f:xc_dist_ig_no_mult}, the distribution of XC values now appears very similar to that of backprop-based XC values, but very different from that of IG-based XC values. By making feature attributions proportional to input feature magnitude, IG is able to generate attribution maps that capture salient features in the input, which is one reason why IG attribution maps makes more sense to a human observer compared to a blurry backprop attribution map (interested readers may visit Sundararajan \emph{et al.} \cite{IG_2017_ICML} for more attribution map examples). As a result, IG zeros out most of the attributions outside of the bounding box (because the space outside of object bounding boxes are mostly empty, leading to zero input feature values at such locations), and the remaining attributions are mostly concentrated within the bounding box or at its close proximity (see \Cref{f:attr_viz}). Thus, IG is unable to produce mostly very low (< 0.1) XC values for the FP predictions, leading to significant overlap in the values for TP prediction XC scores and FP predictions XC scores, making classification using XC scores difficult. 

These observations echo with Erion \emph{et al.}'s \cite{erion2020improving} claim that IG's choice of a zero-valued baseline is problematic. Take an image of a digit for example, if the background is white but the digit itself is black (i.e., zero), then the zero-valued pixels in fact contains the key features of this image and should not get zero attributions. Similarly for point cloud inputs, not just the presence of points, but also the absence of points in certain locations can help the model classify the object. For instance, pedestrian objects are usually filled with points, whereas car objects have points on its boundaries but are mostly hollow in the middle.

Note that even without multiplying by input values, IG-derived XC scores still cannot beat backprop-derived XC scores on the pedestrian and cyclist classes (see \Cref{t:avg_xc_performance}). Hence, a more expensive method such as IG may produce more visually appealing attribution map, but a less expensive method may have more potential when used quantitatively. 

\subsection{Combining XC scores with the top class score}
\label{ss:combine_box_features}

In \Cref{ss:xc_detect_tp_fp} we demonstrate that the top class score is the best performing box-wise feature in classifying TP vs. FP predictions. In this subsection, we aim to improve its performance by combining it with XC scores generated from backprop attributions. We train classifiers for different object classes in KITTI \cite{Geiger2012CVPR} and perform an ablation study on the box-wise features. 

To conduct the ablation study, we build a new dataset $D_f$ from five features (the top classs score 
and the four XC scores computed based on backprop attributions) for each predicted box produced by one PointPillars \cite{Lang_2019_CVPR} model on the KITTI validation set \cite{Geiger2012CVPR}. Since we have three models trained on KITTI, we obtain three $D_f$ datasets.
For each $D_f$, we first group the samples by the predicted object label, then by the number of LiDAR points: those with less than 100 points forms one set, the rest forms another set. Thus, from one $D_f$, we generate six smaller datasets $d_f$, two for each of the three object classes.

We then build a 2-layer multilayer perceptron (MLP) with PyTorch \cite{NEURIPS2019_9015} as a classifier to be trained on $d_f$. Layer 1 is of size ($d \times 3$) and layer 2 is of size ($3 \times 1$), where $d$ represents the number of input features per instance. ReLU activation is applied after the first layer, and the sigmoid function is applied after the second layer to obtain an output score. We train the MLP to classify a predicted box as TP or FP, using binary cross entropy as the loss function. All input features are normalized based on the following equation prior to being fed into the MLP: $z = (x - \mu) / s$, 
where $x$ is the original feature value, $\mu$ is feature mean value, and $s$ is the standard deviation for that feature. We use the Adam optimizer \cite{kingma2017adam} with learning rate set to 0.001. 

For each experiment, we first shuffle $d_f$ and then augment it by duplicating the instances four times. Next we apply 5-fold cross validation to the augmented $d_f$, obtaining 5 different 80\%/20\% train/validation splits. For the training instances, we also add a small uniformly distributed noise $~U(-0.05, +0.05)$ to each feature to help the MLP generalize better. For each different split, we train the MLP for 12 epochs with batch size = 16 and record three evaluation metrics (AUROC, AUPR, and AUPR\_op) of the output score on the validation instances. We repeat the 5-fold cross validation 5 times, obtaining $5 \times 5 = 25$ different values for each of the metrics, and record the average.  Note that the above process is repeated for each $d_f$ in the three $D_f$ we have, each evaluation metric is averaged over the three $D_f$ and shown in \Cref{t:classifier_performance}.

In \Cref{t:classifier_performance}, when the top class score is the only input feature, we evaluate its performance directly and use it as baseline for comparison; when more than one features are used, we evaluate the performance of the MLP output score. The most notable observation is that for predictions containing less than 100 points, combining the XC scores with top class score can often result in better performance in distinguishing TP vs. FP predictions, especially among the pedestrian predictions. The AUPR for pedestrian increased by (0.540 - 0.492) / 0.492 = 9.8\% after combining the 4 XC scores with top class score. The improvement in AUROC for pedestrian predictions and in AUPR for cyclist predictions also exceed 1\%. Among the predictions with more than 100 points, the benefit of incorporating the XC scores is less observable. Note that the top class score alone is already performing very well for these predictions with more points. This might be why it is more difficult to get additional benefit from the XC scores on these predictions. 

\begin{table}[htp]
\caption{Ablation study on the features used to help classify TP vs. FP predictions on KITTI. 
}
\begin{center}
\resizebox{\columnwidth}{!}{%
\begin{tabular}{ c|ccccc|ccc|ccc } 
\toprule
 \multirow{2}{*}{Object Class} & \multicolumn{5}{c}{Features Used} & \multicolumn{3}{c}{Points < 100} & \multicolumn{3}{c}{Points >= 100}\\
 & Top class score & $XC\_c^{-}$ & $XC\_c^{+}$ & $XC\_s^{-}$ & $XC\_s^{+}$ & AUROC & AUPR & AUPR\_op & AUROC & AUPR & AUPR\_op\\
 \midrule
 \multirow{3}{*}{Car} & \checkmark & & & & & 0.958 & 0.926 & 0.977 & 0.956 & 0.980 & 0.914\\
 & \checkmark & \checkmark & & & & 0.958 & 0.927 & 0.977 & 0.950 & 0.978 & 0.903\\
 & \checkmark & \checkmark & \checkmark & \checkmark & \checkmark & 0.960 & 0.927 & 0.979 & 0.957 & 0.980 & 0.918\\
 \midrule
 \multirow{3}{*}{Pedestrian} & \checkmark & & & & & 0.932 & 0.492 & 0.997 & 0.969 & 0.911 & 0.989 \\
 & \checkmark & \checkmark & & & & 0.938 & 0.527 & 0.997 & 0.973 & 0.919 & 0.991 \\
 & \checkmark & \checkmark & \checkmark & \checkmark & \checkmark & 0.944 & 0.540 & 0.997 & 0.973 & 0.920 & 0.992\\
 \midrule
 \multirow{3}{*}{Cyclist} & \checkmark & & & & & 0.958 & 0.765 & 0.996 & 0.982 & 0.947 & 0.995 \\
 & \checkmark & \checkmark & & & & 0.958 & 0.777 & 0.996 & 0.972 & 0.931 & 0.993\\
 & \checkmark & \checkmark & \checkmark & \checkmark & \checkmark & 0.958 & 0.779 & 0.996 & 0.980 & 0.946 & 0.994\\
 \bottomrule
\end{tabular}}
\end{center}
\label{t:classifier_performance}
\end{table}

\section{Conclusion and future work}
\label{s:conc}
To use the explanations quantitatively, we proposed four XC scores to measure the concentration of the attribution values generated for individual predictions. Applying the four XC scores on the task of classifying TP vs. FP predictions led to over 100\% improvement in AUPR on the pedestrian class on both the KITTI and the Waymo datasets compared to simple heuristics such as distance to sensor and number of LiDAR points in bounding box. Although the XC scores alone could not outperform class score in the TP vs. FP classification task, combining class score with the XC scores using an MLP led to notable improvement compared to using class score alone on the pedestrian predictions. 
Thus, it is worthwhile to explore the XC scores further for more use cases such as using it in loss functions to improve model performance, or use it as a tool for adversarial or out-of-distribution sample detection. 

We also discovered that the XC metrics derived from backprop attributions often outperform those derived from IG attributions on TP vs. FP classification for the pedestrian and cyclist objects. This indicates that explanations that are more understandable to a human observer, such as IG, may not offer superior value when analyzed quantitatively and researchers should not discard simpler explanation methods when exploring quantitative use cases for XAI.


\acksection
We would like to thank Huawei Noah’s Ark Lab, Canada, for sponsoring our work. We also thank our colleagues Sean Sedwards, Rick Salay, Chengjie Huang, Harry Nguyen, and Matthew Pitropov for their support and feedback.

{\small
\bibliographystyle{ieee_fullname}
\bibliography{egbib}
}

\end{document}